\documentclass[10pt,twocolumn,letterpaper]{article}

\usepackage{cvpr}
\usepackage{times}
\usepackage{helvet}
\usepackage{courier}
\usepackage{url}
\usepackage{graphicx}
\usepackage{booktabs}
\usepackage{amsfonts}
\usepackage{nicefrac}
\usepackage{microtype}
\usepackage{epsfig}
\usepackage[utf8]{inputenc}
\usepackage[T1]{fontenc}
\usepackage[tight,footnotesize]{subfigure}
\usepackage{amsmath}
\usepackage{amssymb}
\usepackage{amsthm}
\usepackage{bm}
\usepackage{multirow}
\usepackage{makecell}
\usepackage{footmisc}
\usepackage{marvosym}
\usepackage{latexsym}

\usepackage[breaklinks=true,colorlinks,bookmarks=false]{hyperref}

\cvprfinalcopy

\def\cvprPaperID{1855}

\begin{document}

\title{Learning to Transfer Examples for Partial Domain Adaptation}

\author{
Zhangjie Cao$^1$\thanks{Equal contribution, in alphabetic order}, Kaichao You$^1$\footnotemark[1], Mingsheng Long$^1$(\Letter), Jianmin Wang$^1$, and Qiang Yang$^2$\\
$^1$KLiss, MOE; BNRist; School of Software, Tsinghua University, China\\
$^1$Research Center for Big Data, Tsinghua University, China\\
$^1$Beijing Key Laboratory for Industrial Big Data System and Application\\
$^2$Hong Kong University of Science and Technology, China\\
{\small\tt \{caozhangjie14, youkaichao\}@gmail.com, \{mingsheng, jimwang\}@tsinghua.edu.cn}\\
}

\maketitle

\begin{abstract}
Domain adaptation is critical for learning in new and unseen environments. With domain adversarial training, deep networks can learn disentangled and transferable features that effectively diminish the dataset shift between the source and target domains for knowledge transfer. In the era of Big Data, large-scale labeled datasets are readily available, stimulating the interest in partial domain adaptation (PDA), which transfers a recognizer from a large labeled domain to a small unlabeled domain. It extends standard domain adaptation to the scenario where target labels are only a subset of source labels.  Under the condition that target labels are unknown, the key challenges of PDA are how to transfer relevant examples in the shared classes to promote positive transfer and how to ignore irrelevant ones in the source domain to mitigate negative transfer. In this work, we propose a unified approach to PDA, Example Transfer Network (ETN), which jointly learns domain-invariant representations across domains and a progressive weighting scheme to quantify the transferability of source examples. A thorough evaluation on several benchmark datasets shows that ETN consistently achieves state-of-the-art results for various partial domain adaptation tasks.
\end{abstract}

\section{Introduction}
Deep neural networks have significantly advanced the state-of-the-art performance for various machine learning problems \cite{cite:NIPS14AdversarialNet,cite:CVPR16DRL} and applications \cite{cite:CVPR14RCNN,cite:CVPR15FCN,cite:NIPS15FastRCNN}. A common prerequisite of deep neural networks is the rich labeled data to train a high-capacity model to have sufficient generalization power. Such rich supervision is often prohibitive in real-world applications due to the huge cost of data annotation. Thus, to reduce the labeling cost, there is a strong need to develop versatile algorithms that can leverage rich labeled data from a related source domain. However, this domain adaptation paradigm is hindered by the dataset shift underlying different domains, which forms a major bottleneck to adapting the category models to novel target tasks \cite{cite:TKDE10TLSurvey,cite:CVPR11Bias}.

\begin{figure}[tbp]
  \centering
    \includegraphics[width=0.45\textwidth]{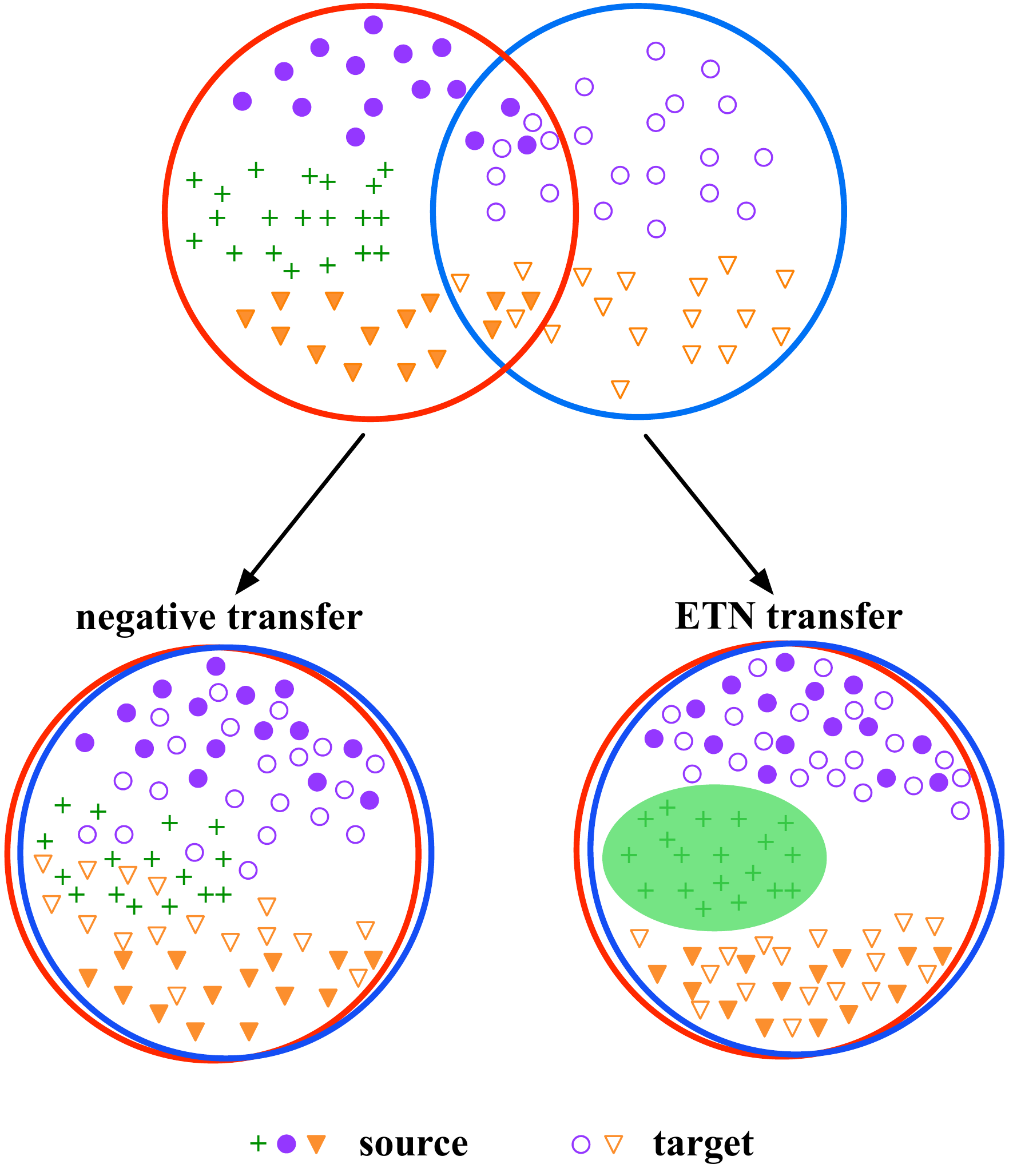}
    \caption{Partial domain adaptation (PDA) is a generalized setting of domain adaptation where the source label space subsumes the target label space. The technical challenge of PDA lies in an intrinsic negative transfer caused by the outlier source classes (`+' in this case), which cannot be forcefully transferred to the target domain. The proposed Example Transfer Network (ETN) designs a weighting scheme to quantify the transferability of source examples and only transfer source examples relevant to the target domain (purple circle and orange triangle), eliminating outlier source examples (in green shadow). Source and target domains are denoted by red and blue circles respectively. \emph{Best viewed in color.}}
    \label{fig:challenge}
\end{figure}

A major line of the existing domain adaptation methods bridge different domains by learning domain-invariant feature representations in the absence of target labels, \ie, unsupervised domain adaptation. Existing methods assume that the source and target domains share the same set of class labels \cite{cite:ECCV10Office,cite:CVPR12GFK}, which is crucial for directly applying the source-trained classifier to the target domain. Recent studies in deep learning reveal that deep networks can disentangle explanatory factors of variations behind domains \cite{cite:ICML14DeCAF,cite:NIPS14CNN}, thus learning more transferable features to improve domain adaptation significantly. These deep domain adaptation methods typically embed distribution matching modules, including moment matching~\cite{cite:Arxiv14DDC,cite:ICML15DAN,cite:NIPS16RTN,cite:ICML17JAN} and adversarial training~\cite{cite:ICML15RevGrad,cite:ICCV15SDT,cite:CVPR17ADDA,cite:NIPS17LuoZHL,cite:ICML18CyCADA}, into deep architectures for end-to-end learning of transferable representations.

Although existing methods can reduce the feature-level domain shift, they assume label spaces across domains are identical. In real-world applications, it is often formidable to find a relevant dataset with the label space identical to the target dataset of interest which is often unlabeled. A more practical scenario is {Partial Domain Adaptation} (PDA)~\cite{cite:CVPR18SAN,cite:CVPR18IWAN,cite:ECCV18PADA}, which assumes that the source label space is a superspace of the target label space, relaxing the constraint of identical label spaces. PDA enables knowledge transfer from a big domain of many labels to a small domain of few labels. With the emergence of Big Data, large-scale labeled datasets such as ImageNet-1K~\cite{cite:ILSVRC15} and Google Open Images~\cite{cite:openimages} are readily accessible to empower data-driven artificial intelligence. These repositories are almost universal to subsume categories of the target domain, making PDA feasible to many applications. PDA can also work in the regime where target data are in limited categories. For example, functions of protein are limited. A large database of known protein structures can be collected, which includes all functions. For a new species, proteins have different structures, but their functions are contained in the database. Predicting protein functions for new species falls into the PDA problem.

As a generalization to standard domain adaptation, partial domain adaptation is more challenging: the target labels are unknown at training, and there must be many ``outlier'' source classes that are useless for the target task. This technical challenge is intuitively illustrated in Figure~\ref{fig:challenge}, where the target classes (like purple `$\circ$' and orange `$\triangledown$') will be forcefully aligned to the outlier source classes (like `+')  by existing domain adaptation methods. As a result, negative transfer will happen because the learner migrates harmful knowledge from the source domain to the target domain. Negative transfer is the principal obstacle to the application of domain adaptation techniques~\cite{cite:TKDE10TLSurvey}. 

Thus, matching the whole source and target domains as previous methods \cite{cite:ICML15DAN,cite:ICML15RevGrad} is not a safe solution to the PDA problem. We need to develop algorithms versatile enough to transfer useful examples from the many-class dataset (source domain) to the few-class dataset (target domain) while robust enough to irrelevant or outlier examples. Three approaches to partial domain adaptation~\cite{cite:CVPR18SAN,cite:CVPR18IWAN,cite:ECCV18PADA} address the PDA by weighing each data point in the domain-adversarial networks, where a domain discriminator is learned to distinguish the source and target. While decreasing the impact of irrelevant examples on domain alignment, they do not undo the negative effect of the outlier classes on the source classifier. Moreover, they evaluate the transferability of source samples without considering the underlying discriminative and multimodal structures. As a result, it is still vulnerable that they may align the features of outlier source classes and target classes, giving way to negative transfer.

Towards a safe approach to partial domain adaptation, we present the Example Transfer Network (ETN), which improves the previous work~\cite{cite:CVPR18SAN,cite:CVPR18IWAN,cite:ECCV18PADA} by learning to transfer useful examples. 
ETN automatically evaluates the transferability of source examples with a transferability quantifier based on their similarities to the target domain, which is used to weigh their contributions to both the source classifier and the domain discriminator. In particular, ETN improves the weight quality over previous work \cite{cite:CVPR18IWAN} by further revealing the discriminative structure to the transferability quantifier. By this means, irrelevant source examples can be better detected and filtered out. Another key improvement of ETN over the previous methods is the capability to simultaneously confine the source classifier and the domain-adversarial network within the auto-discovered shared label space, thus promoting the positive transfer of relevant examples and mitigating negative transfer of irrelevant examples. Comprehensive experiments demonstrate that our model achieves state-of-the-art results on several benchmark datasets, including Office-31, Office-Home, ImageNet-1K, and Caltech-256.

\section{Related Work}

\noindent \textbf{Domain Adaptation}
Domain adaptation, a special scenario of transfer learning \cite{cite:TKDE10TLSurvey}, bridges domains of different distributions to mitigate the burden of annotating target data for machine learning \cite{cite:TNN11TCA,cite:TPAMI12DTMKL,cite:ICML13TCS,cite:NIPS14FTL}, computer vision \cite{cite:ECCV10Office,cite:CVPR12GFK,cite:NIPS14LSDA} and natural language processing \cite{cite:JMLR11MTLNLP}. The main technical difficulty of domain adaptation is to formally reduce the distribution discrepancy across different domains. Deep networks can learn representations that suppress explanatory factors of variations behind data \cite{cite:TPAMI13DLSurvey} and manifest invariant factors across different populations. These invariant factors enable knowledge transfer across relevant domains \cite{cite:NIPS14CNN}. Deep networks have been extensively explored for domain adaptation \cite{cite:CVPR13MidLevel,cite:NIPS14LSDA}, yielding significant performance gains against shallow domain adaptation methods.

While deep representations can disentangle complex data distributions, recent advances show that they can only reduce, but not remove, the cross-domain discrepancy \cite{cite:Arxiv14DDC}. Thus deep learning alone cannot bound the generalization risk for the target task \cite{cite:COLT09DAT,cite:ML10DAT}. Recent works bridge deep learning and domain adaptation \cite{cite:Arxiv14DDC,cite:ICML15DAN,cite:ICML15RevGrad,cite:ICCV15SDT,cite:NIPS16RTN}. They extend deep networks to domain adaptation by adding adaptation layers through which high-order statistics of distributions are explicitly matched \cite{cite:Arxiv14DDC,cite:ICML15DAN,cite:NIPS16RTN}, or by adding a domain discriminator to distinguish features of the source and target domains, while the features are learned adversarially to deceive the discriminator in a minimax game \cite{cite:ICML15RevGrad,cite:ICCV15SDT}.

\noindent \textbf{Partial Domain Adaptation}
While the standard domain adaptation advances rapidly, it still needs the vanilla assumption that the source and target domains share the same label space. This assumption does not hold in partial domain adaptation (PDA), which transfers models from many-class domains to few-class domains. There are three valuable efforts towards the PDA problem. Selective Adversarial Network (SAN)~\cite{cite:CVPR18SAN} adopts multiple adversarial networks with a weighting mechanism to select out source examples in the outlier classes. Partial Adversarial Domain Adaptation \cite{cite:ECCV18PADA} improves SAN by employing only one adversarial network and further adding the class-level weight to the source classifier. Importance Weighted Adversarial Nets (IWAN)~\cite{cite:CVPR18IWAN} uses the Sigmoid output of an auxiliary domain classifier (not involved in domain-adversarial training) to derive the probability of a source example belonging to the target domain, which is used to weigh source examples in the domain-adversarial network. These pioneering approaches achieve dramatical performance gains over standard methods in partial domain adaptation tasks.

These efforts mitigate negative transfer caused by outlier source classes and promote positive transfer among shared classes. However, as outlier classes are only selected out for the domain discriminators, the source classifier is still trained with all classes \cite{cite:CVPR18SAN}, whose performance for shared classes may be distracted by outlier classes. Further, the domain discriminator of IWAN \cite{cite:CVPR18IWAN} for obtaining the importance weights distinguishes the source and target domains only based on the feature representations, without exploiting the discriminative information in the source domain. This will result in non-discriminative importance weights to distinguish shared classes from outlier classes.
This paper proposes an Example Transfer Network (ETN) that down-weights the irrelevant examples of outlier classes further on the source classifier and adopts a discriminative domain discriminator to quantify the example transferability.

\noindent \textbf{Open-Set Domain Adaptation}
On par with domain adaptation, research has been dedicated to open set recognition, with the goal to reject outliers while correctly recognizing inliers during testing. Open Set SVM \cite{cite:ECCV14OSSVM} trains a probabilistic SVM and rejects unknown samples by a threshold. Open Set Neural Network \cite{cite:CVPR16OSNN} generalizes deep neural networks to open set recognition by introducing an OpenMax layer, which estimates the probability of an input from an unknown class and rejects the unknown point by a threshold. Open Set Domain Adaptation (OSDA) \cite{cite:ICCV17OSDA,cite:ECCV18OSBP} tackles the setting when the training and testing data are from different distributions and label spaces. OSDA methods often assume which classes are shared by the source and target domains are known at training. Unlike OSDA, in our scenario, target classes are entirely unknown at training. It is interesting to extend our work to the open set scenario under the generic assumption that all target classes are unknown.

\section{Example Transfer Network}
The scenario of partial domain adaptation (PDA) \cite{cite:CVPR18SAN} constitutes a {source} domain $\mathcal{D}_s = \{(\mathbf{x}_i^s,{\bf y}^s_i)\}_{i=1}^{n_s}$ of $n_s$ labeled examples associated with $|\mathcal{C}_s|$ classes and a {target} domain ${{\mathcal D}_t} = \{ {\mathbf{x}}_j^t\} _{j = 1}^{{n_t}}$ of $n_t$ unlabeled examples drawn from $|\mathcal{C}_t|$ classes. Note that in PDA the source domain label space $\mathcal{C}_s$ is a superspace of the target domain label space $\mathcal{C}_t$ i.e. $\mathcal{C}_s \supset \mathcal{C}_t$. The source and target domains are drawn from different probability distributions $p$ and $q$ respectively. Besides $p \ne q$ as in standard domain adaptation, we further have $p_{\mathcal{C}_t} \ne q$ in partial domain adaptation, where $p_{\mathcal{C}_t}$ denotes the distribution of the source domain data in label space $\mathcal{C}_t$.
The goal of PDA is to learn a deep network that enables end-to-end training of a transferable feature extractor $G_f$ and an adaptive classifier $G_y$ to sufficiently close the distribution discrepancy across domains and bound the target risk ${\Pr _{\left( {{\mathbf{x}},y} \right) \sim q}}\left[ {G_y \left( G_f({\mathbf{x}}) \right) \ne {\bf y}} \right]$.

We incur deteriorated performance when directly applying the source classifier $G_y$ trained with standard domain adaptation methods to the target domain.
In partial domain adaptation, it is difficult to identify which part of the source label space $\mathcal{C}_s$ is shared with the target label space $\mathcal{C}_t$ because the target domain is fully unlabeled and $\mathcal{C}_t$ is unknown at the training stage.
Under this condition, most of existing deep domain adaptation methods \cite{cite:ICML15DAN,cite:ICML15RevGrad,cite:ICCV15SDT,cite:NIPS16RTN} are prone to \emph{negative transfer}, a degenerated case where the classifier with adaptation performs even worse than the classifier without adaptation.
The negative transfer happens since they assume that the source and target domains have identical label space and match whole distributions $p$ and $q$ even though $p_{\mathcal{C}_s \backslash \mathcal{C}_t}$ and $q$ are non-overlapping and cannot be matched in principle.
Thus, decreasing the negative effect of the source examples in outlier label space $\mathcal{C}_s \backslash \mathcal{C}_t$ is the key to mitigating negative transfer in partial domain adaptation.
Besides, we also need to reduce the distribution shift across $p_{\mathcal{C}_t}$ and $q$ to enhance positive transfer in the shared label space $\mathcal{C}_t$ as before. Note that the irrelevant source examples may come from both outlier classes and shared classes, thus requiring a versatile algorithm to identify them.

\begin{figure*}[tbp]
  \centering
  \includegraphics[width=0.9\textwidth]{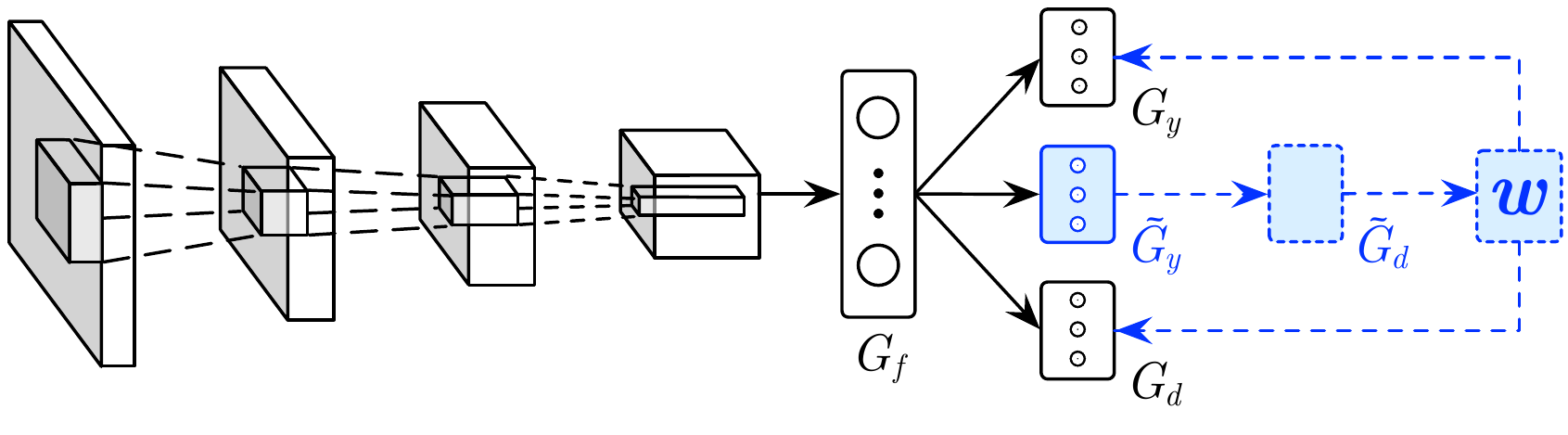}
  \caption{Example Transfer Network (ETN) for partial domain adaptation, where $G_f$ is the feature extractor, $G_y$ is the source classifier, $G_d$ is domain discriminator (involved in adversarial training) for domain alignment; $\tilde G_d$ is the auxiliary domain discriminator (uninvolved in adversarial training) that quantifies the transferability $\mathbf{w}$ of each source example, and $\tilde G_y$ is the auxiliary label predictor encoding the discriminative information to the auxiliary domain discriminator $\tilde G_d$. Modules in blue are newly designed in this paper. \emph{Best viewed in color.}}
   \label{fig:arch}
\end{figure*}

\subsection{Transferability Weighting Framework}
The key technical problem of domain adaptation is to reduce the distribution shift between the source and target domains. Domain adversarial networks~\cite{cite:ICML15RevGrad,cite:ICCV15SDT} tackle this problem by learning transferable features in a two-player minimax game: the first player is a domain discriminator $G_d$ trained to distinguish the feature representations of the source domain from the target domain, and the second player is a feature extractor $G_f$ trained simultaneously to deceive the domain discriminator.

Specifically, the domain-invariant features $\mathbf{f}$ are learned in a minimax optimization procedure: the parameters $\theta_f$ of the feature extractor $G_f$ are trained by maximizing the loss of domain discriminator $G_d$, while the parameters $\theta_d$ of the domain discriminator $G_d$ are trained by minimizing the loss of the domain discriminator $G_d$. Note that our goal is to learn a source classifier that transfers to the target, hence the loss of the source classifier $G_y$ is also minimized. This leads to the optimization problem proposed in~\cite{cite:ICML15RevGrad}:
\begin{equation}\label{eqn:GRL}
\begin{small}
    \begin{aligned}
    E \left( {{\theta _f},{\theta _y},{\theta _d}} \right) & =  \frac{1}{{{n_s}}}\sum\limits_{{{\mathbf{x}}_i} \in {\mathcal{D}_s}} {{L_y}\left( {{G_y}\left( {{G_f}\left( {{{\mathbf{x}}_i}} \right)} \right),{{\bf y}_i}} \right)}  \\
    & - \frac{1}{{n_a}}\sum\limits_{{{\mathbf{x}}_i} \in {\mathcal{D}_a}} {{L_d}\left( {{G_d}\left( {{G_f}\left( {{{\mathbf{x}}_i}} \right)} \right),{{\bf d}_i}} \right)},
    \end{aligned}
    \end{small}
\end{equation}
where $\mathcal{D}_a = \mathcal{D}_s \cup \mathcal{D}_t$ is the union of the source and target domains and $n_a = |\mathcal{D}_a|$, ${\bf d}_i$ is the domain label, $L_y$ and $L_d$ are the cross-entropy loss functions.  

While domain adversarial networks yield reliable results for standard domain adaptation, they will incur performance degeneration on the partial domain adaptation where $\mathcal{C}_s \supset \mathcal{C}_t$. This degeneration is caused by the outlier classes $\mathcal{C}_s \backslash \mathcal{C}_t$ in the source domain, which are undesirably matched to the target classes $\mathcal{C}_t$. Due to the domain gap, even the source examples in the shared label space $\mathcal{D}_t$ may not transfer well to the target domain. As a consequence, we need to design a new framework for partial domain adaptation.

This paper presents a new transferability weighting framework to address the technical difficulties of partial domain adaptation. Denote by $w (\mathbf{x}_i^s)$ the weight of each source example $\mathbf{x}_i^s$, which quantifies the example's transferability. Then for a source example with a larger weight, we should increase its contribution to the final model to enhance positive transfer; otherwise, we should decrease its contribution to mitigating negative transfer. IWAN~\cite{cite:CVPR18IWAN}, a previous work for partial domain adaptation, reweighs the source examples in the loss of the domain discriminator $G_d$. We further put the weights in the loss of the source classifier $G_y$. This significantly enhances our ability to diminish the irrelevant source examples that deteriorate our final model.

Furthermore, the unknownness of target labels can make the identification of shared classes difficult, making partial domain adaptation more difficult. We thus believe that the exploitation of unlabeled target examples by semi-supervised learning is also indispensable. We make use of the entropy minimization principle~\cite{cite:NIPS05EM}. Let ${\hat {\bf y}}_j^t = G_y(G_f(\mathbf{x}_j^t)) \in \mathbb{R}^{|\mathcal{C}_s|}$, the entropy loss to quantify the uncertainty of a target example's predicted label is $ H\left( {{G_y}\left( {{G_f}\left( {{{\mathbf{x}}_j^t}} \right)} \right)} \right) =  - \sum\nolimits_{c = 1}^{|{\mathcal{C}_s}|} {\hat y_{j,c}^t\log \hat y_{j,c}^t} $.

The transferability weighting framework is shown in Figure~\ref{fig:arch}. By weighting the losses of the source classifier $G_y$ and the domain discriminator $G_d$ using the transferability $w(\mathbf{x}_i^s)$ of each source example, and combining the entropy minimization criterion, we achieve the following objective:
\begin{equation}\label{eqn:WL}
\begin{small}
\begin{aligned}
  {E_{G_y}} =\; & \frac{1}{{{n_s}}}\sum\limits_{i = 1}^{{n_s}} {w\left( {{{\bf{x}}_i^s}} \right)L\left( {{G_y}\left( {{G_f}\left({{\bf{x}}_i^s} \right)}, {\bf{y}}_i^s \right)} \right)} \\
  +\; &  \frac{\gamma }{{{n_t}}}\sum\limits_{j = 1}^{{n_t}} {H\left( {G_y}\left( {{G_f}\left({{\bf{x}}_j^t} \right)} \right) \right)},
\end{aligned}
\end{small}
\end{equation}
\begin{equation}\label{eqn:WD}
\begin{small}
\begin{aligned}
{E_{G_d}} = -\; & \frac{1}{{{n_s}}}\sum\limits_{i = 1}^{{n_s}} {w\left( {{\bf{x}}_i^s} \right)\log \left(G_d\left( {G_f}\left({{\bf{x}}_i^s} \right) \right)\right)} \\
- \; & \frac{1}{{{n_t}}}\sum\limits_{j = 1}^{{n_t}} {\log \left( {1 - G_d\left( {G_f}\left({{\bf{x}}_j^t} \right) \right)} \right)},
\end{aligned}
\end{small}
\end{equation}
where $\gamma$ is a hyper-parameter to trade-off the labeled source examples and unlabeled target examples.

The transferability weighting framework can be trained end-to-end by a minimax optimization procedure as follows, yielding a saddle point solution $(\hat\theta_f, \hat\theta_y, \hat\theta_d)$:
\begin{equation}\label{eqn:param1}
\begin{small}
\begin{aligned}
     (\hat\theta_f, \hat\theta_y) &=  \mathop {\arg\min }\limits_{{\theta _f},{\theta _y}} E_{G_y} - E_{G_d}, \\
     (\hat\theta_d) &=  \mathop {\arg\min }\limits_{{\theta_d}} E_{G_d}.
\end{aligned}
\end{small}
\end{equation}

\subsection{Example Transferability Quantification}

With the proposed transferability weighting framework in Equations~\eqref{eqn:WL} and \eqref{eqn:WD}, the key technical problem is how to quantify the transferability of each source example $w(\mathbf{x}_i^s)$. We introduce an auxiliary domain discriminator $\tilde G_d$, which is also trained to distinguish the representations of the source domain from the target domain, using the similar loss as Equation~\eqref{eqn:WD} but dropping $w(\mathbf{x}_i^s)$. It is not involved in the adversarial training procedure, \ie, the features $G_f$ are not learned to confuse $\tilde G_d$. Such an auxiliary domain discriminator can roughly quantify the transferability of the source examples, through the Sigmoid probability of classifying each source example $\mathbf{x}_i^s$ to the target domain.



Such an auxiliary domain discriminator ${\tilde G_d}$ discriminates source and target domains based on the assumption that source examples of shared classes $\mathcal{C}_t$ are closer to the target domain than to those source examples in the outlier classes $\mathcal{C}_s \backslash \mathcal{C}_t$, thus having higher probability to be predicted as from the target domain.
However, the auxiliary domain discriminator only distinguishes the source and target examples based on domain information. There is potential small gap between ${\tilde G_d}$'s outputs for transferable and irrelevant source examples especially when ${\tilde G_d}$ is trained well. So the model is still exposed to the risk of mixing up the transferable and irrelevant source examples, yielding unsatisfactory transferability measures $w(\mathbf{x}_i^s)$. In partial domain adaptation, the source examples in $\mathcal{C}_t$ differentiate from those in $\mathcal{C}_s \backslash \mathcal{C}_t$ mainly in that $\mathcal{C}_t$ is shared with the target domain while $\mathcal{C}_s \backslash \mathcal{C}_t$ has no overlap with the target domain. Thus, it is natural to integrate discriminative information into our weight design to resolve the ambiguity between shared and outlier classes.

Inspired by AC-GANs \cite{cite:ICML17ACGAN} that integrate the labeled information into the discriminator, we aim to integrate the label information into the auxiliary domain discriminator $\tilde G_d$. However, we hope to develop a transferability measure $w(\mathbf{x}_i^s)$ with both the discriminative information and domain information to generate clearly separable weights for source data in $\mathcal{C}_t$ and $\mathcal{C}_s \backslash \mathcal{C}_t$ respectively. Thus, we add an auxiliary label predictor $\tilde G_y$ with leaky-softmax activation. Within $\tilde G_y$, the feature from feature extractor $G_f$ are transformed to $|{\cal C}_s|$-dimension $\mathbf{z}$. Then $\mathbf{z}$ will be passed through a leaky-softmax activation as follows,
\begin{equation}
 \tilde\sigma\left( \mathbf{z} \right) = \frac{{\exp \left( \mathbf{z} \right)}}{{|{\cal C}_s| + \sum\nolimits_{c = 1}^{|{\cal C}_s|} {\exp \left( {{z_c}} \right)} }},
\end{equation}
where $z_c$ is the $c$-th dimension of $\mathbf{z}$. The leaky-softmax has the property that the element-sum of its outputs is smaller than $1$; when the logit $z_c$ of class $c$ is very large, the probability to classify an example as class $c$ is high. As the auxiliary label predictor $\tilde G_y$ is trained on source examples and labels, the source examples will have higher probability to be classified as a specific source class $c$, while the target examples will have smaller logits and uncertain predictions. 
Therefore, the element-sum of the leaky-softmax outputs are closer to $1$ for source examples and closer to $0$ for target examples. If we define $\tilde G_d$ as
\begin{equation}\label{eqn:aux_domain}
  \tilde G_d \left( {{G_f}\left( {{{\mathbf{x}}_i}} \right)} \right) = \sum\nolimits_{c = 1}^{|{\mathcal{C}_s}|} {{\tilde G_y^c}\left( {{G_f}\left( {{{\mathbf{x}}_i}} \right)} \right)} ,
\end{equation}
where $\tilde G_y^c\left( {G_f}\left( {{{\mathbf{x}}_i}} \right)\right)$ is the probability of each example $\mathbf{x}_i$ belonging to class $c$, then $\tilde G_d \left( {{G_f}\left( {{{\mathbf{x}}_i}} \right)} \right)$ can be seen as computing the probability of each example belonging to the source domain. For a source example, the smaller the value of $\tilde G_d \left( {{G_f}\left( {{{\mathbf{x}}_i}} \right)} \right)$ is, the more probable that it comes from the target domain, meaning that it is closer to the target domain and more likely to be in the shared label space $\mathcal{C}_t$. Thus, the output of $\tilde G_d$ is suitable for transferability quantification.


We train the auxiliary label predictor $\tilde G_y$ with the leaky-softmax by a multitask loss over $|\mathcal{C}_s|$ one-vs-rest binary classification tasks for the $|\mathcal{C}_s|$-class classification problem:
\begin{equation}\label{eqn:aux_class}
\begin{aligned}
{E_{\tilde G_y}} = & - \frac{\lambda }{{{n_s}}}\sum\limits_{i = 1}^{{n_s}} \sum\limits_{c = 1}^{|\mathcal{C}_s|} \left[ y_{i,c}^s\log {\tilde G_y^c}\left( {G_f}\left({{\bf{x}}_i^s} \right) \right) \right.\\
&\left. + \left( {1 - y_{i,c}^s} \right)\log \left( {1 - {\tilde G_y^c}\left( {G_f}\left({{\bf{x}}_i^s} \right) \right)} \right) \right] ,
\end{aligned}
\end{equation}
where $y_{i,c}^s$ denotes whether class $c$ is the ground-truth label for source example ${\bf{x}}_i^s$, and $\lambda$ is a hyper-parameter. We also train the auxiliary domain discriminator $\tilde G_d$ to distinguish the features of the source domain and the target domain as
\begin{equation}\label{eqn:Dhat}
\begin{aligned}
{E_{\tilde G_d}} = -\; & \frac{1}{{{n_s}}}\sum\limits_{i = 1}^{{n_s}} {\log \left(\tilde G_d\left( {G_f}\left({{\bf{x}}_i^s} \right) \right)\right)} \\
- \; & \frac{1}{{{n_t}}}\sum\limits_{j = 1}^{{n_t}} {\log \left( {1 - \tilde G_d\left( {G_f}\left({{\bf{x}}_j^t} \right) \right)} \right)}.
\end{aligned}
\end{equation}

From Equations~\eqref{eqn:aux_domain} to \eqref{eqn:Dhat}, we observe that the outputs of the auxiliary domain discriminator ${\tilde G_d}$ depend on the outputs of the auxiliary label predictor $\tilde G_y$. This guarantees that ${\tilde G_d}$ is trained with both label and domain information, resolving the ambiguity between shared and outlier classes to better quantify the example transferability.

Finally, with the help of the auxiliary label predictor $\tilde{G}_y$ and the auxiliary domain discriminator $\tilde{G}_d$, we can derive more accurate and discriminative weights to quantify the transferability of each source example as
\begin{equation}\label{eqn:W}
  w\left( {{{\mathbf{x}}_i^s}} \right) = 1 - \tilde G_d \left( {{G_f}\left( {{{\mathbf{x}}_i^s}} \right)} \right).
\end{equation}
Since the outputs of $\tilde G_d$ for source examples are closer to $1$, implying very small weights, we normalize the weights in each mini-batch of batch size $B$ as $w\left( {\bf{x}} \right) \leftarrow \frac{{w\left( {\bf{x}} \right)}}{{\frac{1}{B}\sum\nolimits_{i = 1}^B {w\left( {{{\bf{x}}_i}} \right)} }}$.


\subsection{Minimax Optimization Problem}

With the aforementioned derivation, we now formulate our final model, Example Transfer Network (ETN). We unify the transferability weighting framework in Equations~\eqref{eqn:WL}--\eqref{eqn:WD} and the example transferability quantification in Equations~\eqref{eqn:aux_domain}--\eqref{eqn:W}.
Denoting by $\theta_{\tilde y}$ the parameters of the auxiliary label predictor $\tilde G_y$, the proposed ETN model can be solved by a minimax optimization problem that finds saddle-point solutions $\hat\theta_f$, $\hat\theta_y$, $\hat\theta_d$ and $\hat\theta_{\tilde y}$ to model parameters as follows,
\begin{equation}
\begin{small}
\begin{aligned}
     (\hat\theta_f, \hat\theta_y) & =  \mathop {\arg\min }\limits_{{\theta _f},{\theta _y}} E_{G_y} - E_{G_d}, \\
     (\hat\theta_d) & =  \mathop {\arg\min }\limits_{{\theta_d}} E_{G_d}, \\
     (\hat\theta_{\tilde y}) & =  \mathop {\arg\min }\limits_{{\theta_{\tilde y}}} {E_{\tilde G_y}} + {E_{\tilde G_d}}.
\end{aligned}
\end{small}
\end{equation}

ETN enhances partial domain adaptation by learning to transfer relevant examples and diminish outlier examples for both source classifier $G_y$ and domain discriminator $G_d$. It exploits progressive weighting schemes $w(\mathbf{x}_i^s)$ from the auxiliary domain discriminator $\tilde{G}_d$ and auxiliary label predictor $\tilde{G}_y$, well quantifying the transferability of source examples.


\begin{table*}[htbp]
    \addtolength{\tabcolsep}{1.5pt}
    \centering
    \caption{Classification Accuracy (\%) for Partial Domain Adaptation on {Office-Home} Dataset (\textbf{ResNet-50})}
    \label{table:accuracy_officehome}
    \resizebox{\textwidth}{!}{%
    \begin{tabular}{cccccccccccccc}
        \toprule
        \multirow{2}{30pt}{\centering Method} & \multicolumn{13}{c}{Office-Home} \\
        \cmidrule(lr){2-14}
        & {Ar}$\rightarrow${Cl} & {Ar}$\rightarrow${Pr} & {Ar}$\rightarrow${Rw} & {Cl}$\rightarrow${Ar} & {Cl}$\rightarrow${Pr} & {Cl}$\rightarrow${Rw} & {Pr}$\rightarrow${Ar} & {Pr}$\rightarrow${Cl} & {Pr}$\rightarrow${Rw} & {Rw}$\rightarrow${Ar} & {Rw}$\rightarrow${Cl} & {Rw}$\rightarrow${Pr} & Avg \\
        \midrule
        ResNet~\cite{cite:CVPR16DRL} & 46.33 & 67.51 & 75.87 & 59.14 & 59.94 & 62.73 & 58.22 & 41.79 & 74.88 & 67.40 & 48.18 & 74.17 & 61.35 \\
        DANN~\cite{cite:ICML15RevGrad} & 43.76 & 67.90 & 77.47 & 63.73 & 58.99 & 67.59 & 56.84 & 37.07 & 76.37 & 69.15 & 44.30 & 77.48 & 61.72 \\
        ADDA~\cite{cite:CVPR17ADDA} & 45.23 & 68.79 & 79.21 & 64.56 & 60.01 & 68.29 & 57.56 & 38.89 & 77.45 & 70.28 & 45.23 & 78.32 & 62.82 \\
        RTN~\cite{cite:NIPS16RTN} & 49.31 & 57.70 & \textbf{80.07} & 63.54 & 63.47 & 73.38 & 65.11 & 41.73 & 75.32 & 63.18 & 43.57 & 80.50 & 63.07 \\
        IWAN~\cite{cite:CVPR18IWAN} & 53.94 & 54.45 & 78.12 & 61.31 & 47.95 & 63.32 & 54.17 & 52.02 & 81.28 & \textbf{76.46} & 56.75 & 82.90 & 63.56 \\
        SAN~\cite{cite:CVPR18SAN} & 44.42 & 68.68 & 74.60 & \textbf{67.49} & 64.99 & \textbf{77.80} & 59.78 & 44.72 & 80.07 & 72.18 & 50.21 & 78.66 & 65.30 \\
        PADA~\cite{cite:ECCV18PADA} &51.95 & 67.00 & 78.74 & 52.16 & 53.78 & 59.03 & 52.61 & 43.22 & 78.79 & 73.73 & 56.60 & 77.09 & 62.06 \\
        \midrule
        ETN & \textbf{59.24} & \textbf{77.03} & 79.54 & 62.92 & \textbf{65.73} & 75.01 & \textbf{68.29} & \textbf{55.37} & \textbf{84.37} & 75.72 & \textbf{57.66} & \textbf{84.54} & \textbf{70.45} \\
        \bottomrule
    \end{tabular}%
    }
\end{table*}

\begin{table*}[htbp]
    \addtolength{\tabcolsep}{-1pt}
    \centering
    \caption{Classification Accuracy (\%) for Partial Domain Adaptation on {Office-31} and {ImageNet-Caltech} Datasets (\textbf{ResNet-50})}
    \label{table:accuracy_officeic}
    \resizebox{1\textwidth}{!}{%
    \begin{tabular}{ccccccccccl}
        \toprule
        \multirow{2}{30pt}{\centering Method} &  \multicolumn{7}{c}{Office-31} & \multicolumn{3}{c}{ImageNet-Caltech} \multirow{2}{50pt}{\centering Avg} \\
        \cmidrule(lr){2-8} \cmidrule(lr){9-10}
        & A$\rightarrow$W & D$\rightarrow$W & W$\rightarrow$D & A$\rightarrow$D & D$\rightarrow$A & W$\rightarrow$A & Avg & I $\rightarrow$ C & C $\rightarrow$ I &  \\
        \midrule
        ResNet~\cite{cite:CVPR16DRL} & 75.59$\pm$1.09 & 96.27$\pm$0.85 & 98.09$\pm$0.74 & 83.44$\pm$1.12 & 83.92$\pm$0.95 & 84.97$\pm$0.86 & 87.05$\pm$0.94 & 69.69$\pm$0.78 & 71.29$\pm$0.74 & 70.49$\pm$0.76 \\
        DAN~\cite{cite:ICML15DAN} & 59.32$\pm$0.49 & 73.90$\pm$0.38 & 90.45$\pm$0.36 & 61.78$\pm$0.56 & 74.95$\pm$0.67 & 67.64$\pm$0.29 & 71.34$\pm$0.46 & 71.30$\pm$0.46 & 60.13$\pm$0.50 & 65.72$\pm$0.48 \\
        DANN~\cite{cite:ICML15RevGrad} & 73.56$\pm$0.15 & 96.27$\pm$0.26 & 98.73$\pm$0.20 & 81.53$\pm$0.23 & 82.78$\pm$0.18 & 86.12$\pm$0.15 & 86.50$\pm$0.20 & 70.80$\pm$0.66 & 67.71$\pm$0.76 & 69.23$\pm$0.71 \\
        ADDA~\cite{cite:CVPR17ADDA} & 75.67$\pm$ 0.17 & 95.38$\pm$0.23 & 99.85$\pm$0.12 & 83.41$\pm$ 0.17 & 83.62$\pm$0.14 & 84.25$\pm$0.13 & 87.03$\pm$0.16 & 71.82$\pm$0.45 & 69.32$\pm$0.41 & 70.57$\pm$0.43 \\
        RTN~\cite{cite:NIPS16RTN} & 78.98$\pm$0.55 & 93.22$\pm$0.52 & 85.35$\pm$0.47 & 77.07$\pm$0.49 & 89.25$\pm$0.39 & 89.46$\pm$0.37 & 85.56$\pm$0.47 & 75.50$\pm$0.29 & 66.21$\pm$0.31 & 70.85$\pm$0.30 \\
        IWAN~\cite{cite:CVPR18IWAN} & 89.15$\pm$0.37 & 99.32$\pm$0.32 & 99.36$\pm$0.24 & 90.45$\pm$0.36 & 95.62$\pm$0.29 & 94.26$\pm$0.25 & 94.69$\pm$0.31 & 78.06$\pm$0.40 & 73.33$\pm$0.46 & 75.70$\pm$0.43 \\
        SAN~\cite{cite:CVPR18SAN} & 93.90$\pm$0.45 & 99.32$\pm$0.52 & 99.36$\pm$0.12 & 94.27$\pm$0.28 & 94.15$\pm$0.36 & 88.73$\pm$0.44 & 94.96$\pm$0.36 & 77.75$\pm$0.36 & \textbf{75.26}$\pm$0.42 & 76.51$\pm$0.39 \\
        PADA~\cite{cite:ECCV18PADA} &86.54$\pm$0.31 & 99.32$\pm$0.45 & \textbf{100.00}$\pm$.00 & 82.17$\pm$0.37 & 92.69$\pm$0.29 & \textbf{95.41}$\pm$0.33 & 92.69$\pm$0.29 & 75.03$\pm$0.36 & 70.48$\pm$0.44 & 72.76$\pm$0.40 \\
        \midrule
        ETN & \textbf{94.52}$\pm$0.20 & \textbf{100.00}$\pm$.00 & \textbf{100.00}$\pm$.00 & \textbf{95.03}$\pm$0.22 & \textbf{96.21}$\pm$0.27 & 94.64$\pm$0.24 & \textbf{96.73}$\pm$0.16 & \textbf{83.23}$\pm$0.24 & 74.93$\pm$0.28 & \textbf{79.08}$\pm$0.26 \\
        \bottomrule
    \end{tabular}%
    }
\end{table*}

\begin{table*}[!htbp]
    \addtolength{\tabcolsep}{3pt}
    \centering
    \caption{Classification Accuracy (\%) for Partial Domain Adaptation on {Office-31} Dataset (\textbf{VGG})}
    \label{table:accuracy_officevgg}
    \resizebox{0.85\textwidth}{!}{%
    \begin{tabular}{cccccccc}
        \toprule
        \multirow{2}{30pt}{\centering Method} & \multicolumn{7}{c}{Office-31} \\
        \cmidrule(lr){2-8}
        & A$\rightarrow$W & D$\rightarrow$W & W$\rightarrow$D & A$\rightarrow$D & D$\rightarrow$A & W$\rightarrow$A & Avg \\
        \midrule
        VGG~\cite{cite:ICLR15VGG} & 60.34$\pm$0.84 & 97.97$\pm$0.63 & 99.36$\pm$0.36 & 76.43$\pm$0.48 & 72.96$\pm$0.56 & 79.12$\pm$0.54 & 81.03$\pm$ 0.57 \\
        DAN~\cite{cite:ICML15DAN} & 58.78$\pm$0.43 & 85.86$\pm$0.32 & 92.78$\pm$0.28 & 54.76$\pm$0.44 & 55.42$\pm$0.56 & 67.29$\pm$0.20 & 69.15$\pm$0.37 \\
        DANN~\cite{cite:ICML15RevGrad} & 50.85$\pm$0.12 & 95.23$\pm$0.24 & 94.27$\pm$0.16 & 57.96$\pm$0.20 & 51.77$\pm$0.14 & 62.32$\pm$0.12 & 68.73$\pm$0.16 \\
        ADDA~\cite{cite:CVPR17ADDA} & 53.28$\pm$0.15 & 94.33$\pm$0.18 & 95.36$\pm$0.08 & 58.78$\pm$0.12 & 50.24$\pm$0.10 & 63.34$\pm$0.08 & 69.22$\pm$0.12 \\
        RTN~\cite{cite:NIPS16RTN} & 69.35$\pm$0.42 & 98.42$\pm$0.48 & 99.59$\pm$0.32 & 75.43$\pm$0.38 & 81.45$\pm$0.32 & 82.98$\pm$0.36 & 84.54$\pm$0.38 \\
        IWAN~\cite{cite:CVPR18IWAN} & 82.90$\pm$0.31 & 79.75$\pm$0.26 & 88.53$\pm$0.16 & \textbf{90.95}$\pm$0.33 & 89.57$\pm$0.24 & 93.36$\pm$0.22 & 87.51$\pm$0.25 \\
        SAN~\cite{cite:CVPR18SAN} & 83.39$\pm$0.36 & 99.32$\pm$0.45 & \textbf{100.00}$\pm$.00 & 90.70$\pm$0.20 & 87.16$\pm$0.23 & 91.85$\pm$0.35 & 92.07$\pm$0.27 \\
        PADA~\cite{cite:ECCV18PADA} & \textbf{86.05}$\pm$0.36 & 99.42$\pm$0.24 & \textbf{100.00}$\pm$.00 & 81.73$\pm$0.34 & 93.00$\pm$0.24 & \textbf{95.26}$\pm$0.27 & 92.54$\pm$0.24 \\
        \midrule
        ETN & 85.66$\pm$0.16 & \textbf{100.00}$\pm$.00 & \textbf{100.00}$\pm$.00 & 89.43$\pm$0.17 & \textbf{95.93}$\pm$0.23 & 92.28$\pm$0.20 & \textbf{96.74}$\pm$0.13 \\
        \bottomrule
    \end{tabular}%
    }
\end{table*}

\begin{table*}[!htbp]
    \addtolength{\tabcolsep}{0pt}
    \centering
    \caption{Classification Accuracy (\%) of ETN and Its Variants for Partial Domain Adaptation on {Office-Home} Dataset (\textbf{ResNet-50})}
    \label{table:accuracy_ablation}
    \resizebox{\textwidth}{!}{%
    \begin{tabular}{cccccccccccccc}
        \toprule
        \multirow{2}{30pt}{\centering Method} & \multicolumn{13}{c}{Office-Home} \\
        \cmidrule(lr){2-14}
        & {Ar}$\rightarrow${Cl} & {Ar}$\rightarrow${Pr} & {Ar}$\rightarrow${Rw} & {Cl}$\rightarrow${Ar} & {Cl}$\rightarrow${Pr} & {Cl}$\rightarrow${Rw} & {Pr}$\rightarrow${Ar} & {Pr}$\rightarrow${Cl} & {Pr}$\rightarrow${Rw} & {Rw}$\rightarrow${Ar} & {Rw}$\rightarrow${Cl} & {Rw}$\rightarrow${Pr} & Avg \\
        \midrule
        ETN w/o classifier & 56.18 & 71.93 & 79.32 & \textbf{65.11} & 65.57 & 73.66 & 65.47 & 52.90 & 82.88 & 72.93 & 56.93 & 82.91 & 68.93 \\
        ETN w/o auxiliary & 48.36 & 50.42 & 79.13 & 56.57 & 45.88 & 65.49 & 56.38 & 49.07 & 77.53 & 75.57 & \textbf{58.81} & 78.32 & 61.79 \\
        ETN & \textbf{59.24} & \textbf{77.03} & \textbf{79.54} & 62.92 & \textbf{65.73} & \textbf{75.01} & \textbf{68.29} & \textbf{55.37} & \textbf{84.37} & \textbf{75.72} & 57.66 & \textbf{84.54} & \textbf{70.45} \\
        \bottomrule
    \end{tabular}%
    }
    \vspace{-10pt}
\end{table*}

\section{Experiments}
We conduct experiments to evaluate our approach with state-of-the-art (partial) domain adaptation methods. Codes and datasets will be available at \url{github.com/thuml}.

\subsection{Setup}


\noindent \textbf{Office-31}~\cite{cite:ECCV10Office} is \textit{de facto} for domain adaptation. It is relatively small with 4,652 images in 31 classes. Three domains, namely \textbf{A, D, W},  are collected by downloading from \textbf{amazon.com} (\textbf{A}), taking from \textbf{DSLR} (\textbf{D}) and from \textbf{web camera} (\textbf{W}). Following the protocol in \cite{cite:CVPR18SAN}, we select images from the 10 categories shared by \textbf{Office-31} and \textbf{Caltech-256} to build new target domain, creating six partial domain adaptation tasks: \textbf{A}$\rightarrow$\textbf{W}, \textbf{D}$\rightarrow$\textbf{W}, \textbf{W}$\rightarrow$\textbf{D}, \textbf{A}$\rightarrow$\textbf{D}, \textbf{D}$\rightarrow$\textbf{A} and \textbf{W}$\rightarrow$\textbf{A}. Note that there are 31 categories in the source domain and 10 categories in the target domain.

\noindent \textbf{Office-Home}~\cite{cite:CVPR17OfficeHome} is a larger dataset, with 4 domains of distinct styles: \textbf{Artistic}, \textbf{Clip Art}, \textbf{Product} and \textbf{Real-World}. Each domain contains images of 65 object categories. Denoting them as \textbf{Ar}, \textbf{Cl}, \textbf{Pr}, \textbf{Rw}, we obtain twelve partial domain adaptation tasks: \textbf{Ar}$\rightarrow$\textbf{Cl}, \textbf{Ar}$\rightarrow$\textbf{Pr}, \textbf{Ar}$\rightarrow$\textbf{Rw}, \textbf{Cl}$\rightarrow$\textbf{Ar}, \textbf{Cl}$\rightarrow$\textbf{Pr}, \textbf{Cl}$\rightarrow$\textbf{Rw}, \textbf{Pr}$\rightarrow$\textbf{Ar}, \textbf{Pr}$\rightarrow$\textbf{Cl}, \textbf{Pr}$\rightarrow$\textbf{Rw}, \textbf{Rw}$\rightarrow$\textbf{Ar}, \textbf{Rw}$\rightarrow$\textbf{Cl}, and \textbf{Rw}$\rightarrow$\textbf{Pr}. For PDA, we use images from the first 25 classes in alphabetical order as the target domain and images from all 65 classes as the source domain.

\noindent \textbf{ImageNet-Caltech} is a large dataset built with \textbf{ImageNet-1K}~\cite{cite:ILSVRC15} and \textbf{Caltech-256}. They share 84 classes, and thus we form two partial domain adaptation tasks: \textbf{ImageNet (1000)}$\rightarrow$\textbf{Caltech (84)} and \textbf{Caltech (256)}$\rightarrow$\textbf{ImageNet (84)}. As most networks are trained on the training set of ImageNet, we use images from ImageNet validation set as target domain for \textbf{Caltech (256)}$\rightarrow$\textbf{ImageNet (84)} task.

We compare the proposed \textbf{ETN} model with state-of-the-art deep learning and (partial) domain adaptation methods: \textbf{ResNet-50}~\cite{cite:CVPR16DRL}, Deep Adaptation
Network (\textbf{DAN})~\cite{cite:ICML15DAN}, Domain-Adversarial Neural Networks (\textbf{DANN})~\cite{cite:ICML15RevGrad}, Adversarial Discriminative Domain Adaptation (\textbf{ADDA})~\cite{cite:CVPR17ADDA}, Residual
Transfer Networks (\textbf{RTN})~\cite{cite:NIPS16RTN}, Selective Adversarial Network (\textbf{SAN})~\cite{cite:CVPR18SAN}, Importance Weighted Adversarial Network (\textbf{IWAN})~\cite{cite:CVPR18IWAN} and Partial Adversarial Domain Adaptation (\textbf{PADA})~\cite{cite:ECCV18PADA}. 

Besides ResNet-50 \cite{cite:CVPR16DRL}, we also evaluate ETN and some methods based on \textbf{VGG}~\cite{cite:ICLR15VGG} on the {Office-31} dataset.
We perform ablation study to justify the example transfer mechanism, by evaluating two \textbf{ETN} variants: 1) \textbf{ETN w/o classifier} is the variant without weights on the source classifier; 2) \textbf{ETN w/o auxiliary} is the variant without the auxiliary label predictor on the auxiliary domain discriminator.

We implement all methods based on \textbf{PyTorch}, and fine-tune ResNet-50~\cite{cite:CVPR16DRL} and VGG~\cite{cite:ICLR15VGG} pre-trained on ImageNet. New layers are trained from scratch, and their learning rates are 10 times that of the fine-tuned layers. We use mini-batch SGD with momentum of 0.9 and the learning rate decay strategy implemented in DANN~\cite{cite:ICML15RevGrad}: the learning rate is adjusted during SGD using $\eta_p = \frac{\eta_0}{{(1+\alpha p)}^\beta}$, where $p$ is the training progress linearly changing from $0$ to $1$. The flip-coefficient of the gradient reversal layer is increased gradually from $0$ to $1$ as DANN \cite{cite:ICML15RevGrad}. Hyper-parameters are optimized with importance weighted cross-validation~\cite{cite:JMLR07IWCV}.

\begin{figure*}[htbp]
    \centering
    \subfigure[DANN]{
        \includegraphics[width=0.2\textwidth]{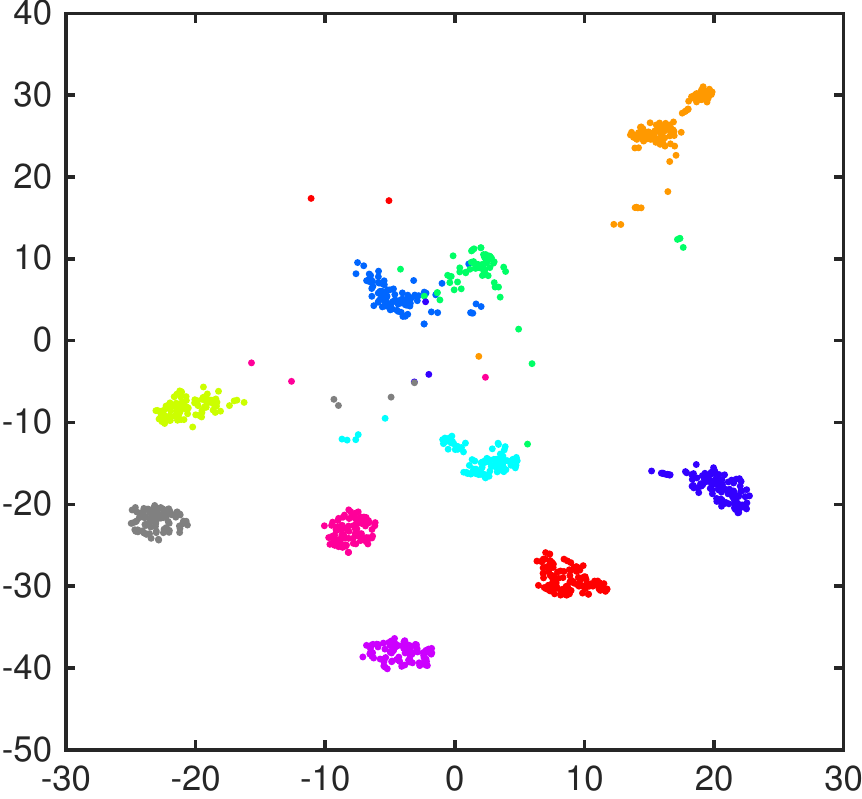}
        \label{fig:dann}
    }\hfil
    \subfigure[SAN]{
        \includegraphics[width=0.2\textwidth]{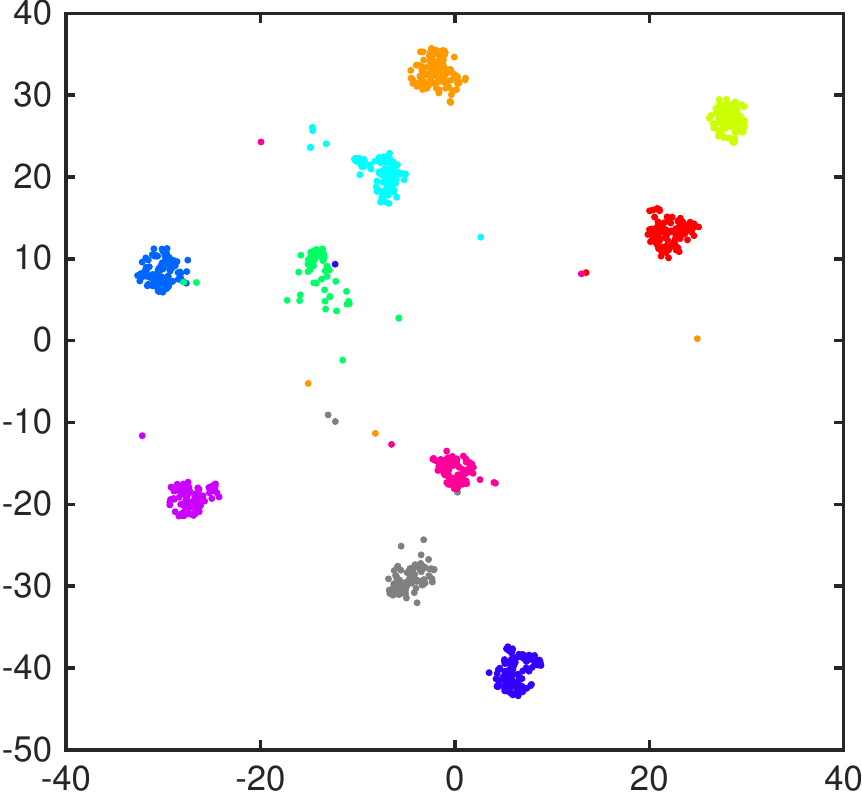}
        \label{fig:san}
    }\hfil
    \subfigure[IWAN]{
        \includegraphics[width=0.2\textwidth]{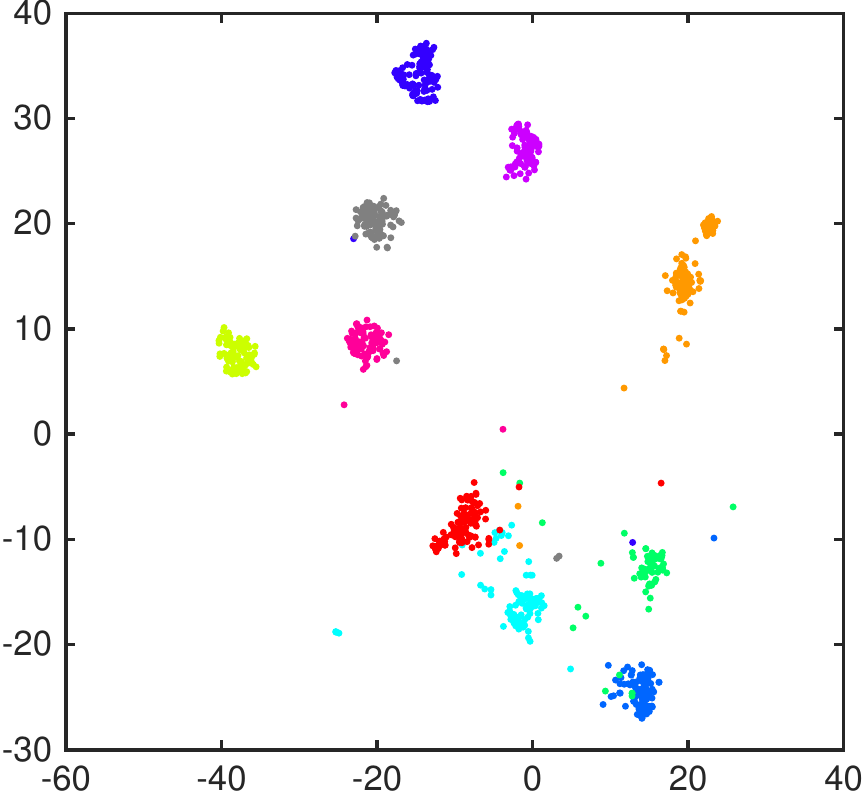}
        \label{fig:iwan}
    }\hfil
    \subfigure[ETN]{
        \includegraphics[width=0.2\textwidth]{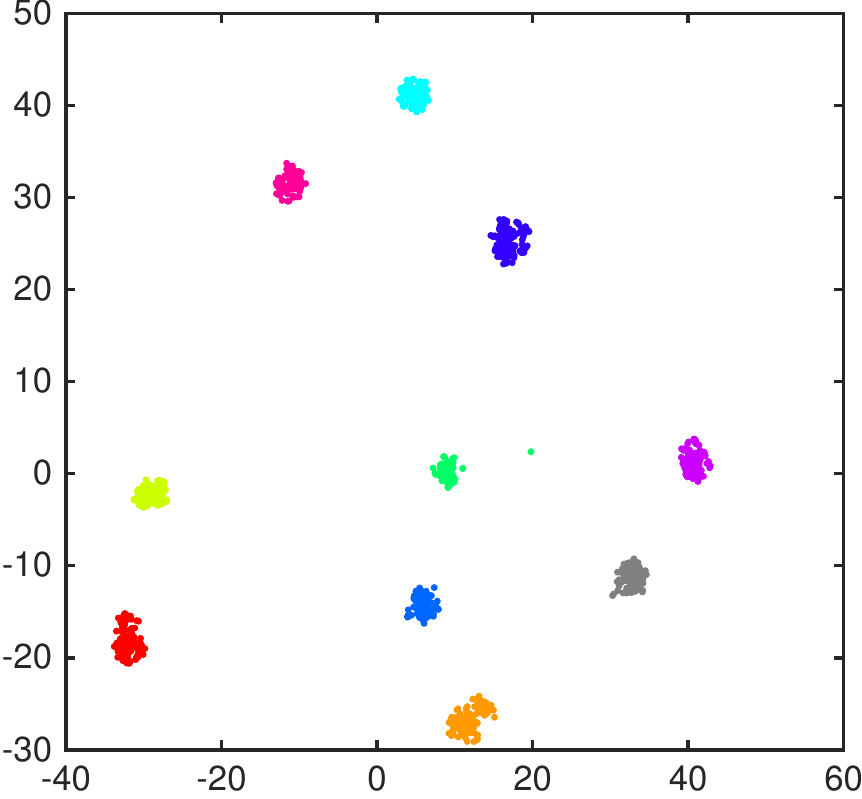}
        \label{fig:ETN}
    }
    \caption{Visualization of features learned by DANN, SAN, IWAN, and ETN (class information is denoted by different colors).}
    \label{fig:tnse}
    \vspace{-5pt}
\end{figure*}

\begin{figure}[htbp]
    \centering
    \includegraphics[width=0.6\columnwidth]{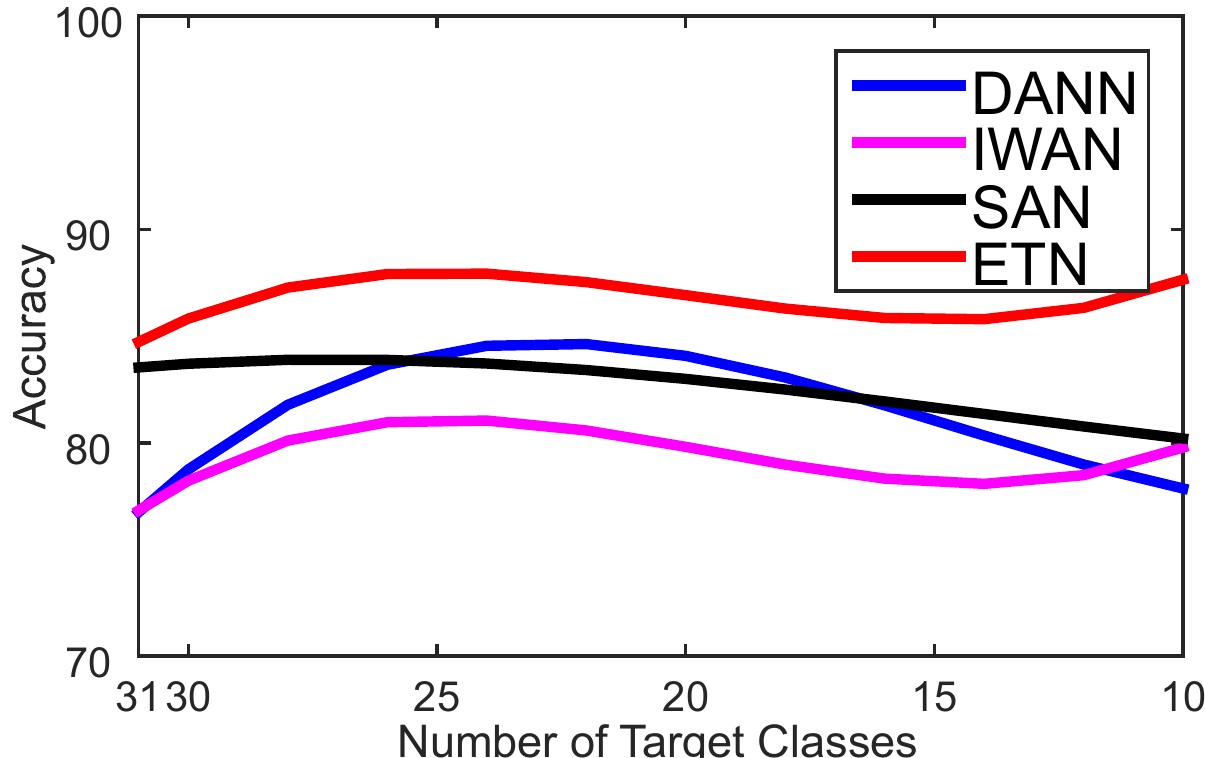}
    \caption{Accuracy by varying \#target classes.}
    \label{fig:accuracy_number}
\end{figure}

\begin{figure}[htbp]
    \centering
    \includegraphics[width=0.6\columnwidth]{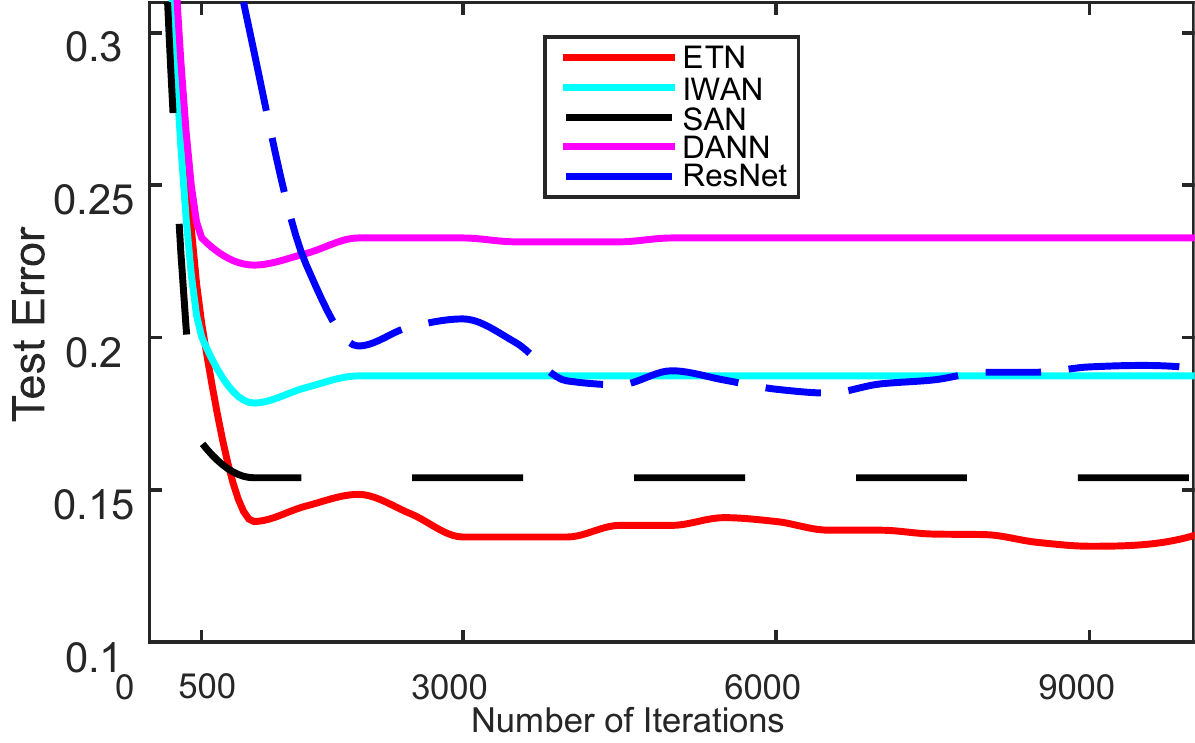}
    \caption{Target test error w.r.t. to \#iterations.}
    \label{fig:validation_error}
\end{figure}

\subsection{Results}
The classification results based on \textbf{ResNet-50} on the the twelve tasks of \textbf{Office-Home}, six tasks of \textbf{Office-31} and the two large-scale tasks of \textbf{ImageNet-Caltech} are shown in Tables \ref{table:accuracy_officehome} and \ref{table:accuracy_officeic}. We also compare all methods on \textbf{Office-31} with \textbf{VGG} backbone in Table~\ref{table:accuracy_officevgg}. ETN outperforms all other methods w.r.t average accuracy, showing that ETN performs well with different base networks on different datasets.

Specifically, we have several observations. \textbf{1)} ADDA, DANN, and DAN outperform ResNet only on some tasks, implying that they suffer from the negative transfer issue.
\textbf{2)} RTN exploits the entropy minimization criterion to amend itself with semi-supervised learning. Thus, it has some improvement over ResNet but still suffers from negative transfer for some tasks.
\textbf{3)} Partial domain adaptation methods (SAN~\cite{cite:CVPR18SAN} and IWAN~\cite{cite:CVPR18IWAN}) perform better than ResNet and other domain adaptation methods on most tasks, due to their weighting mechanism to mitigate negative transfer caused by outlier classes and promote positive transfer among shared classes.
\textbf{4)} ETN outperforms SAN and IWAN on most tasks, showing its power to discriminate the outlier classes from the shared classes accurately and to transfer relevant examples.

In particular, ETN outperforms SAN and IWAN by much larger margin on the large-scale \textbf{ImageNet-Caltech} dataset, indicating that ETN is robuster to outlier classes and performs better even on dataset with large number of outlier classes ($916$ in \textbf{ImageNet}$\rightarrow$\textbf{Caltech}) relative to the shared classes ($84$ in \textbf{ImageNet}$\rightarrow$\textbf{Caltech}). ETN has two advantages: learning discriminative weights and filtering outlier classes out from both source classifier and domain discriminator, which boost partial domain adaptation performance.

We inspect the efficacy of different modules by comparing in Tables~\ref{table:accuracy_ablation} the results of ETN variants. \textbf{1)} ETN outperforms ETN w/o classifier, proving that the weighting mechanism on the source classifier can reduce the negative influence of outlier-classes examples and focus the source classifier on the examples belonging to the target label space. \textbf{2)} ETN also outperforms ETN w/o auxiliary by a larger margin, proving that the auxiliary classifier can inject label information into the domain discriminator to yield discriminative weights, which in turn enables ETN to filter out irrelevant examples.

\subsection{Analysis}

\noindent \textbf{Feature Visualization:} We plot in Figures~\ref{fig:tnse} the t-SNE embeddings~\cite{cite:ICML14DeCAF} of the features learned by DANN, SAN, IWAN and ETN on \textbf{A (31 classes)}$\rightarrow$\textbf{W (10 classes)} with class information in the target domain. We observe that features learned by DANN, IWAN, and SAN are not clustered as clearly as ETN, indicating that ETN can better discriminate target examples than the compared methods.

\noindent \textbf{Class Overlap:}
We conduct a wide range of partial domain adaptation with different numbers of target classes. Figure~\ref{fig:accuracy_number} shows that when the number of target classes decreases fewer than $23$, the performance of DANN degrades quickly, implying that negative transfer becomes severer when the label space overlap becomes smaller. The performance of SAN decreases slowly and stably, indicating that SAN potentially eliminates the influence of outlier classes. IWAN only performs better than DANN when the label space non-overlap is very large and negative transfer is very severe.
ETN performs stably and consistently better than all compared methods, showing the advantage of ETN to partial domain adaptation.
ETN also performs better than DANN in standard domain adaptation when the label spaces totally overlap, implying that the weighting mechanism will not degrade performance when there are no outlier classes.

\begin{figure}[!htbp]
    \centering
    \subfigure[IWAN]{
        \includegraphics[width=0.2\textwidth]{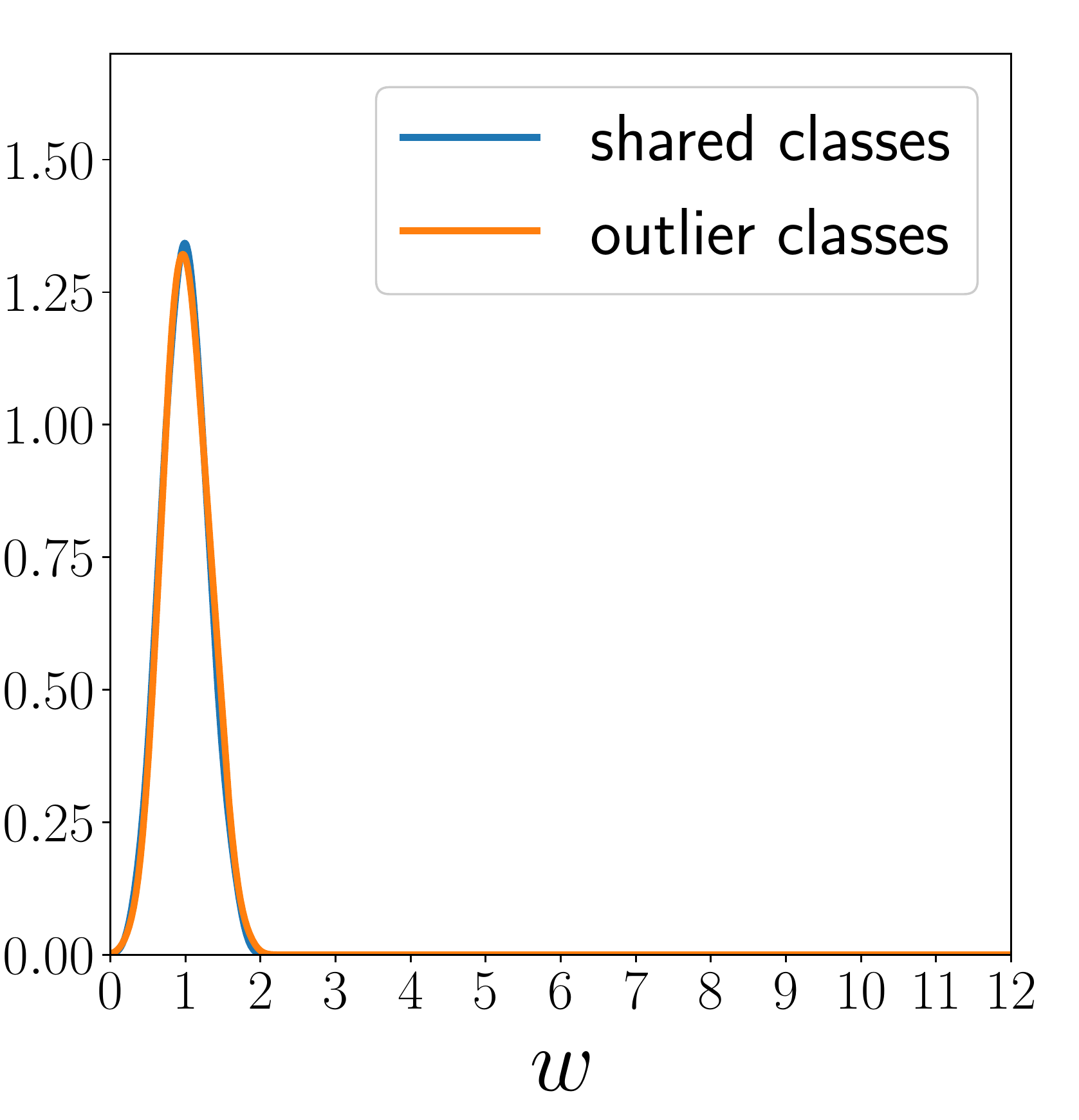}
        \label{fig:IWAN_weight}
    }
    \hfil
    \subfigure[ETN]{
        \includegraphics[width=0.2\textwidth]{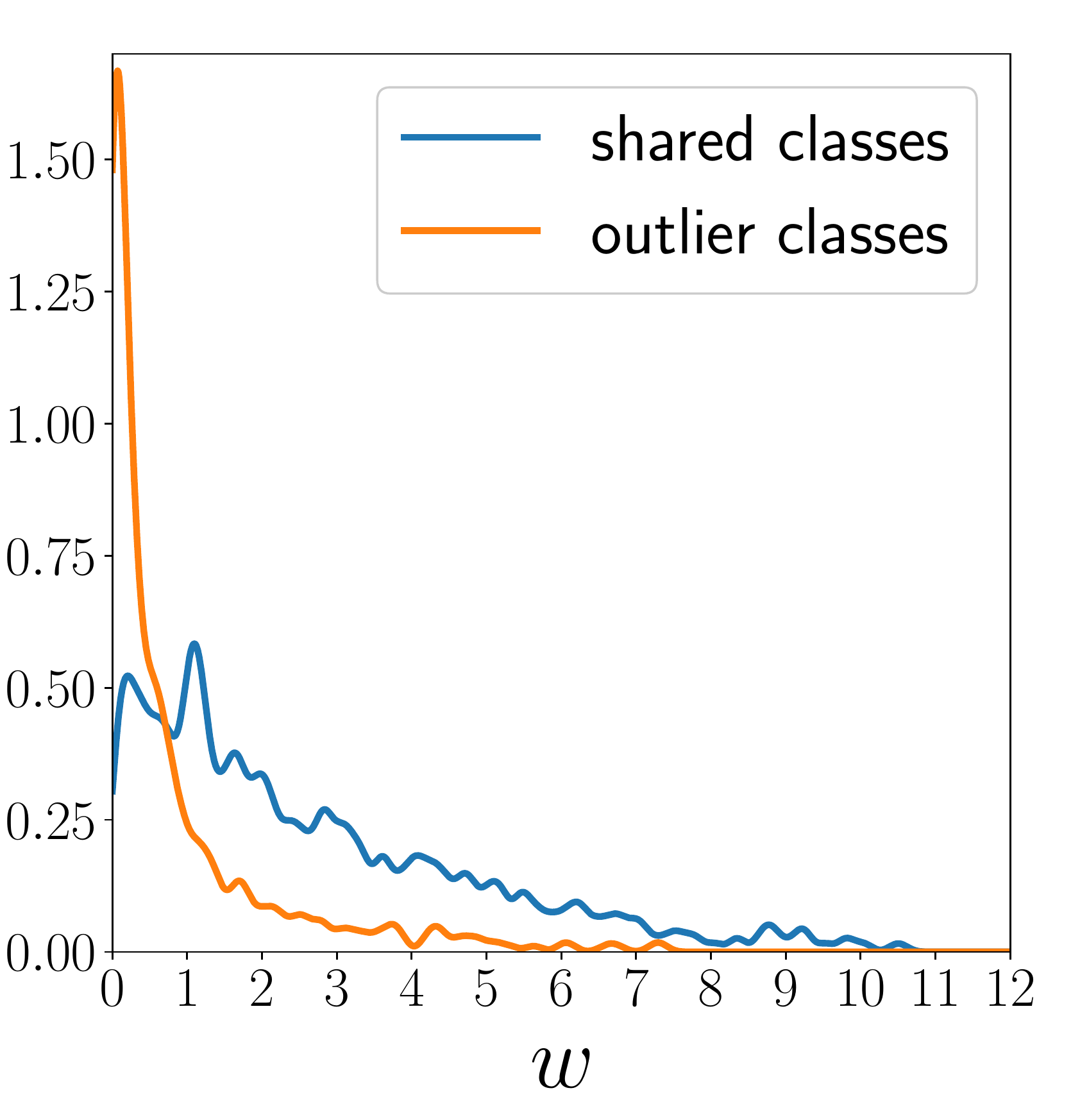}
        \label{fig:ETN_weight}
    }
    \caption{Density function of the importance weights of source examples in the shared label space $\mathcal{C}_t$ and outlier label space $\mathcal{C}_s \backslash \mathcal{C}_t$.}
    \label{fig:weight_distribution}
\end{figure}

\noindent \textbf{Convergence Performance:}
As shown in Figure~\ref{fig:validation_error}, the test errors of all methods converge fast but baselines to high error rates. Only ETN converges to the lowest test error. Such phenomenon implies that ETN can be trained more efficiently and stably than previous domain adaptation methods.

\noindent \textbf{Weight Visualization:} 
We plot the approximate density function of the weights in Equation~\eqref{eqn:W} generated by IWAN \cite{cite:CVPR18IWAN} and ETN for all source examples in Figure~\ref{fig:weight_distribution} on task \textbf{Cl}$\rightarrow$\textbf{Pr}. The orange curve shows examples in shared classes $\mathcal{C}_t$ and the blue curve shows outlier classes $\mathcal{C}_s \backslash \mathcal{C}_t$. Compared to IWAN, our ETN approach assigns much larger weights to shared classes and much smaller weights to outlier classes. Most examples of outlier classes have nearly zero weights, explaining the strong performance of ETN on these datasets.

\section{Conclusion}
This paper presented Example Transfer Network (ETN), a discriminative and robust approach to partial domain adaptation. It quantifies the transferability of source examples by integrating the discriminative information into the transferability quantifier and down-weights the negative influence of the outlier source examples upon both the source classifier and the domain discriminator. Based on the evaluation, our model performs strongly for partial domain adaptation tasks.

\section*{Acknowledgements} 
This work is supported by National Key R\&D Program of China (No.~2016YFB1000701) and National Natural Science Foundation of China (61772299, 71690231, and 61672313).

{\small
\bibliographystyle{ieee_fullname}
\bibliography{main}
}

\end{document}